\documentclass[12pt,a4paper]{article}
\usepackage{epsfig}
\usepackage{icapr}

\title{REFINEMENT OF A STRUCTURED LANGUAGE MODEL\footnote{%
This work was funded by the NSF IRI-19618874 grant STIMULATE}}

\author{%
  Ciprian Chelba, CLSP, The Johns Hopkins University\\
  Frederick Jelinek, CLSP, The Johns Hopkins University\\
}

\summary{%
  A new language model for speech recognition inspired by linguistic analysis is presented. 
  The model develops hidden hierarchical structure incrementally and uses it
  to extract meaningful information from the word history --- thus
  enabling the use of extended distance dependencies --- in an attempt to
  complement the locality of currently used n-gram Markov models.
  The model, its probabilistic parametrization, a reestimation
  algorithm for the model parameters and a set of experiments meant to
  evaluate its potential for speech recognition are presented.
}

\begin{document}
\maketitle
\thispagestyle{empty}

\section{INTRODUCTION}

The task of a speech recognizer is to automatically transcribe speech
into text. The most successful approach to speech recognition so far is a
statistical one~\cite{jelinek83}: given the observed string of acoustic
features $A$, find the most likely word string $\hat{W}$ among those that could
have generated $A$:
\begin{eqnarray}
  \hat{W} = argmax_{W} P(W|A) = argmax_{W} P(A|W)\cdot P(W) \label{intro:bayes}
\end{eqnarray}

This paper is concerned with the estimation of the language model
probability $ P(W)$.
We will first describe current modeling approaches to the problem,
followed by a detailed explanation  of our model. A few preliminary experiments that show the
potential of our approach for language modeling will then be presented.

\section{BASIC LANGUAGE MODELING}

The language modeling problem is to estimate the source probability
$P(W)$ where $W = w_1,w_2,\ldots,w_n$ is a sequence of words. 
This probability is estimated from a text training corpus. Usually the model is
parameterized: $P_{\theta}(W), \theta
\in \Theta$ where $\Theta$ is referred to as the parameter space. 
Due to the sequential nature of an efficient search algorithm, the
model operates left-to-right, allowing the computation
\begin{eqnarray}
P(w_1,w_2,\ldots,w_n) = P(w_1) \cdot \prod_{i=2}^n P(w_i/w_1 \ldots w_{i-1}) \label{eq:l2r}
\end{eqnarray}

We thus seek to develop parametric conditional models:
\begin{eqnarray}
  P_{\theta}(w_i/w_1 \ldots w_{i-1}), \theta \in \Theta, w_i \in \mathcal{V}
\end{eqnarray}
where $\mathcal{V}$ is the vocabulary chosen by the modeler. 
Currently most successful is the \emph{n-gram language model}:
\begin{eqnarray}
  P_{\theta}(w_i/w_{1} \ldots w_{i-1}) = P_{\theta}(w_i/w_{i-n+1} \ldots w_{i-1}) \label{basic_lm:n-gram}
\end{eqnarray}

\subsection{LANGUAGE MODEL QUALITY}

All attempts to derive an algorithm
that would estimate the model parameters so as to minimize the word error rate have
failed. As an alternative, a statistical model is evaluated by
how well it predicts a string of symbols $W_t$ --- commonly named
\emph{test data} --- generated by the source to be modeled.

\subsubsection{Perplexity}

Assume we compare two models $M_1$ and $M_2$; they assign probability
$P_{M_1}(W_t)$ and $P_{M_2}(W_t)$, respectively, to the sample test
string $W_t$. ``Naturally'', we consider $M_1$ to be a better model than $M_2$
if $P_{M_1}(W_t) > P_{M_2}(W_t)$. The test data is not seen during the
model estimation process. 

A commonly used quality measure for a given model $M$ is related to
the entropy of the underlying source and was introduced under
the name of perplexity (PPL)~\cite{jelinek97}:
\begin{eqnarray}
  PPL(M) = exp(-1/|W_t| \sum_{i=1}^{N}\ln{[P_M(W_t)]}) \label{basic_lm:ppl}
\end{eqnarray}

\subsection{SMOOTHING}

Assume that our model $M$
is faced with the prediction $w_i|w_1 \ldots w_{i-1}$ and that $w_i$
has not been seen in the training corpus in context $w_1 \ldots
w_{i-1}$ which itself has possibly not been encountered in the training corpus. 
If $P_M(w_i|w_1 \ldots w_{i-1}) = 0$ then $P_M(w_1 \ldots w_N) = 0$ thus forcing a recognition error;
good models are smooth, in the sense that\\ $\exists
\epsilon(M) > 0$~s.t.~$ P_M(w_i|w_1 \ldots w_{i-1}) > \epsilon,
\forall w_i \in \mathcal{V}$, $(w_1 \ldots w_{i-1}) \in {\mathcal{V}}^{i-1}$.

One standard approach that ensures smoothing is the deleted
interpolation method~\cite{jelinek80}. It interpolates
linearly  among contexts of different order $h_n$:
\begin{eqnarray}
  P_{\theta}(w_i|w_{i-n+1} \ldots w_{i-1}) = \sum_{k=0}^{k=n} \lambda_k \cdot f(w_i/h_k)
\end{eqnarray}
where:
$h_k = w_{i-k+1} \ldots w_{i-1}$ is the context of order $k$
when predicting $w_i$; $f(w_i/h_k)$ is the relative frequency estimate
for the conditional probability $P(w_i/h_k)$; $\lambda_k, k = 0 \ldots n$ are the interpolation
coefficients satisfying $\lambda_k > 0,  k = 0 \ldots n$ and
$\sum_{k=0}^{k=n} \lambda_k = 1$.

The model parameters $\theta$ then are:
the counts $C(h_n,w_i)$ --- lower order counts are inferred recursively
by: $C(h_k,w_i) = \sum_{w_{i-k} \in \mathcal{V}} C(w_{i-k},h_k,w_i)$
---  and the interpolation coefficients $\lambda_k, k = 0 \ldots n$.

A simple way to estimate the model parameters involves a two stage process:
\begin{enumerate}
  \item gather counts from \emph{development data} --- about 90\% of
    training data;
  \item estimate interpolation coefficients to minimize the perplexity
    of \emph{check data} --- the remaining 10\% of the training data.
\end{enumerate}

Different smoothing techniques are also used e.g., maximum
entropy~\cite{berger:max_ent} or back-off~\cite{katz:back_off}.

\section{DESCRIPTION OF THE STRUCTURED LANGUAGE MODEL}

The model we present is closely related to the one
investigated in~\cite{ws96}, however different in a few important
aspects:
\begin{itemize}
\item our model operates in a left-to-right manner, thus allowing its
  use directly in the hypothesis search for $\hat{W}$ in (\ref{intro:bayes});
\item our model is a factored version of the one in~\cite{ws96}, thus enabling the
  calculation of the joint probability of words and parse structure;
  this was not possible in the previous case due to the huge
  computational complexity of the model.
\end{itemize}

\subsection{THE BASIC IDEA AND TERMINOLOGY} \label{section:basic_idea}

\begin{figure}[ha]
  \begin{center} 
    \epsfig{file=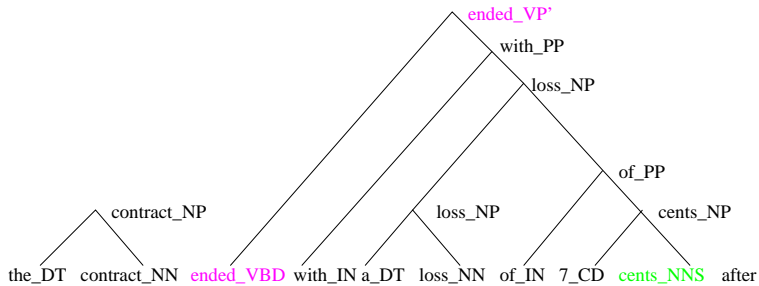,width=10cm}
  \end{center}
  \caption{Partial parse} \label{fig:p_parse}
\end{figure}

Consider predicting the word \verb+after+ in the sentence:\\
\verb+the contract ended with a loss of 7 cents after trading as low as 89 cents+.\\
A 3-gram approach would predict \verb+after+ from \verb+(7, cents)+
whereas it is intuitively clear that the strongest word-pair predictor would be
\verb+contract ended+ which is outside the reach of even 7-grams.
Our assumption is that what enables humans to make a
good prediction of \verb+after+ is the syntactic structure of its sentence
prefix. The linguistically correct \emph{partial parse} of
this prefix is shown in Figure~\ref{fig:p_parse}. 

A binary branching parse for a string of words is a binary tree whose leaves
are the words. The headword annotation makes the tree an
oriented graph: at each node we have two children; the current node
receives a \emph{headword} from either child; one arrow suffices to
describe which of the children --- left or right --- is
percolated to become the headword of the parent.
It was found that better parse trees are generated when using
tags: part-of-speech(POS) tags for the
leaves and non-terminal(NT) tags for the intermediate nodes in the parse
tree. Any subtree identifies a \emph{constituent}. 
The word \verb+ended+ is called the \emph{headword} of the
\emph{constituent} \verb+(ended (with (...)))+
and \verb+ended+ is an \emph{exposed headword} when predicting
\verb+after+ --- topmost headword in the largest constituent that
contains it. The syntactic structure in the past filters out
irrelevant words and points to the important ones, thus enabling the
use of long distance information when predicting the next word. 

Our model will attempt to build the syntactic structure incrementally
while traversing the sentence left-to-right; it will assign a
probability $P(W,T)$ to every sentence $W$ with every possible POStag
assignment, binary branching parse, non-terminal tag and headword
annotation for every constituent of the parse tree $T$. 

\begin{figure}[ha]
  \begin{center}
    \epsfig{file=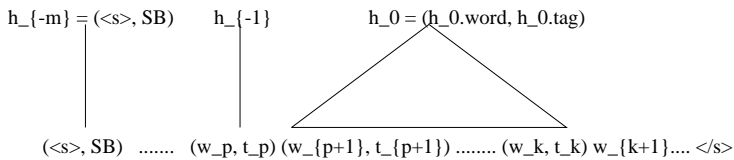,width=10cm}
  \end{center}
  \caption{A word-parse k-prefix} \label{fig:w_parse}
\end{figure}

Let $W$ be a sentence of length $n$ words to which we have prepended
\verb+<s>+ and appended \verb+</s>+ so that $w_0 = $\verb+<s>+ and
$w_{n+1} = $\verb+</s>+.
Let $W_k$ be the word k-prefix $w_0 \ldots w_k$ of the sentence and 
\mbox{$W_k T_k$} the \emph{word-parse k-prefix}. A
\mbox{word-parse k-prefix} contains --- for a given parse --- only those binary
subtrees whose span is completely included in the word k-prefix, excluding 
$w_0 = $\verb+<s>+. Single words along with their POStag can be
regarded as root-only subtrees. Figure~\ref{fig:w_parse} shows a
word-parse k-prefix; \verb|h_0 .. h_{-m}| are the \emph{exposed
 heads}, each head being a pair (headword, non-terminal tag), or
(word,  POStag) in the case of a root-only tree. 

\begin{figure}[ha]
  \begin{center} 
    \epsfig{file=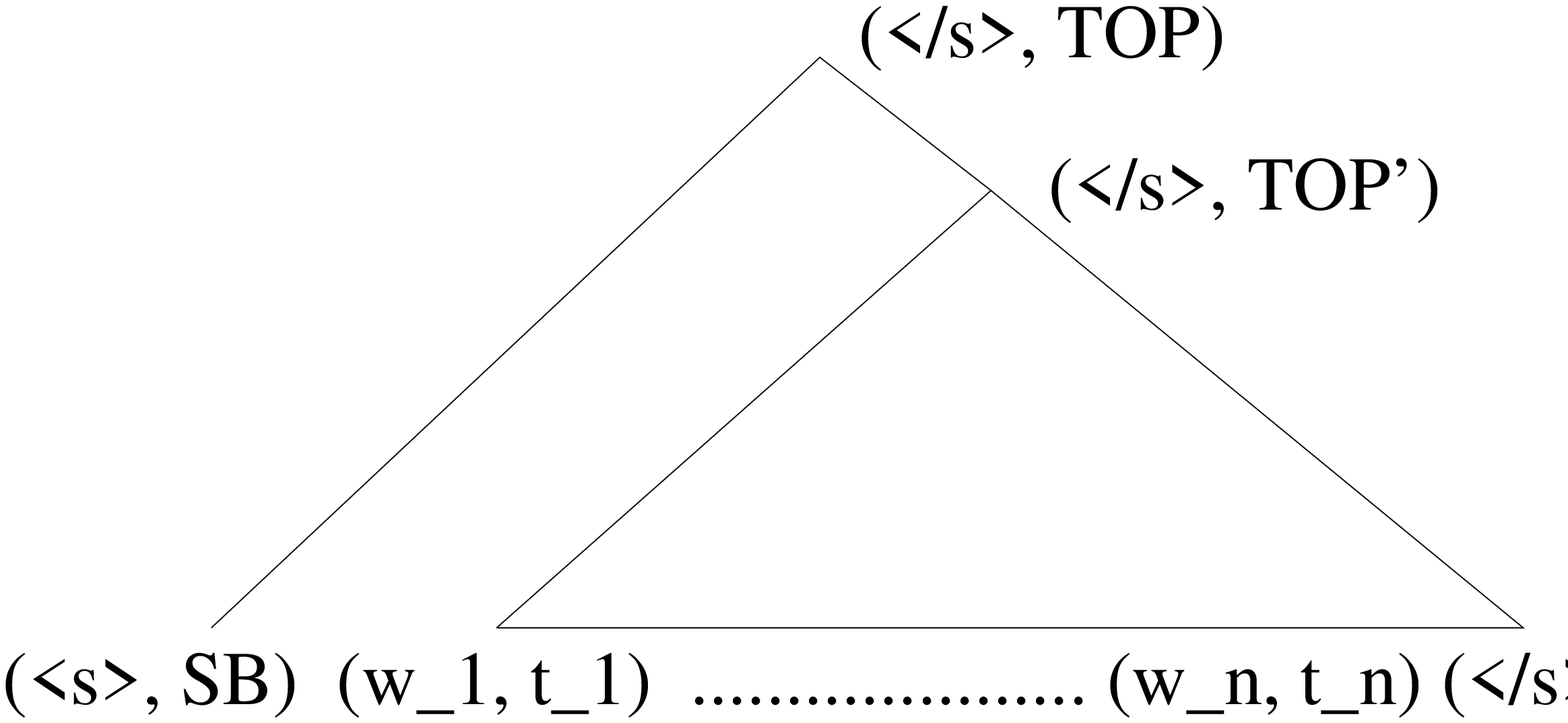,width=6cm}
  \end{center}
  \caption{Complete parse} \label{fig:c_parse}
\end{figure}

A \emph{complete parse} --- Figure~\ref{fig:c_parse} --- is a binary
parse of the \\ \verb+(<s>, SB)+ $(w_1,t_1)  \ldots (w_n,t_n)$\verb+ (</s>, SE)+
sequence with the following two restrictions:
\begin{enumerate} 
\item $(w_1,t_1)  \ldots (w_n,t_n)$\verb+ (</s>, SE)+ is a
  constituent, headed by \verb+(</s>, TOP')+;
\item \verb+(</s>, TOP)+ is the only allowed head. Note that 
  \mbox{$((w_1,t_1) \ldots (w_n,t_n))$} \emph{needn't} be a constituent,
  but for the parses where it is, there is no restriction on which of
  its words is the headword or what is the non-terminal tag that
  accompanies the headword. 
\end{enumerate}
Our model can generate all and only the complete parses for a string\\
\verb+(<s>, SB)+ $(w_1,t_1)  \ldots (w_n,t_n)$\verb+ (</s>, SE)+. 

The model will operate by means of three modules:
\begin{itemize}
\item WORD-PREDICTOR predicts the next word $w_{k+1}$ given the
  word-parse k-prefix \mbox{$W_k T_k$} and then passes control to the TAGGER;
\item TAGGER predicts the POStag $t_{k+1}$ of the next word given the
  word-parse k-prefix and the newly predicted word $w_{k+1}$ and then passes
  control to the PARSER;
\item PARSER grows the already existing binary branching structure by
  repeatedly generating the transitions:
  \verb+(adjoin-left, NTtag)+ or  \verb+(adjoin-right, NTtag)+
  until it passes control to the PREDICTOR
  by  taking a \verb+null+ transition. \verb+NTtag+ is the non-terminal
  tag assigned to each newly built constituent and
  \verb+{left,right}+ specifies from where the new headword is
  inherited. The parser operates always on the two rightmost exposed
  heads, starting with the newly tagged word $w_{k+1}$.
\end{itemize}

\begin{figure}[h]
  \begin{center} 
    \epsfig{file=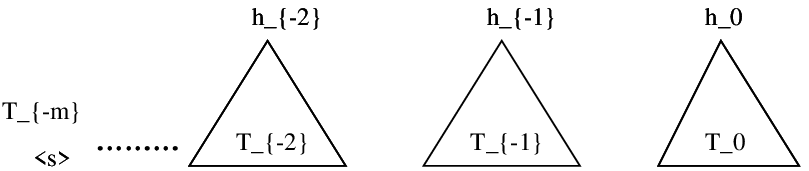,width=8cm}
    \caption{Before an adjoin operation} \label{fig:before}  
    \vspace{0.1cm}
    \epsfig{file=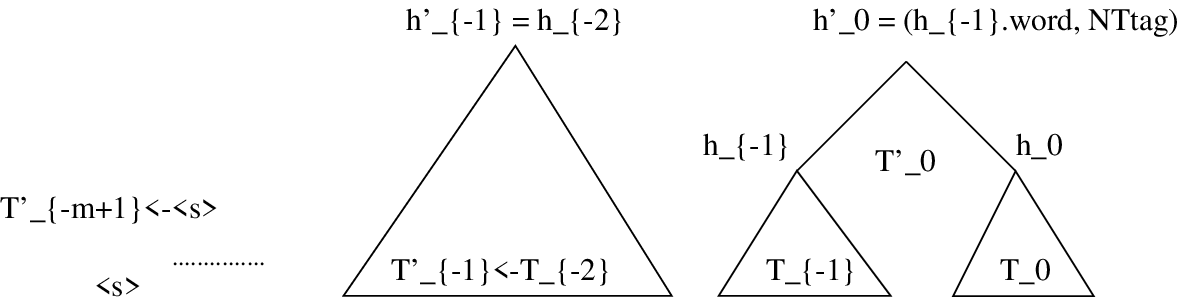,width=8cm}
    \caption{Result of adjoin-left under NTtag} \label{fig:after_a_l}
    \vspace{0.1cm}
    \epsfig{file=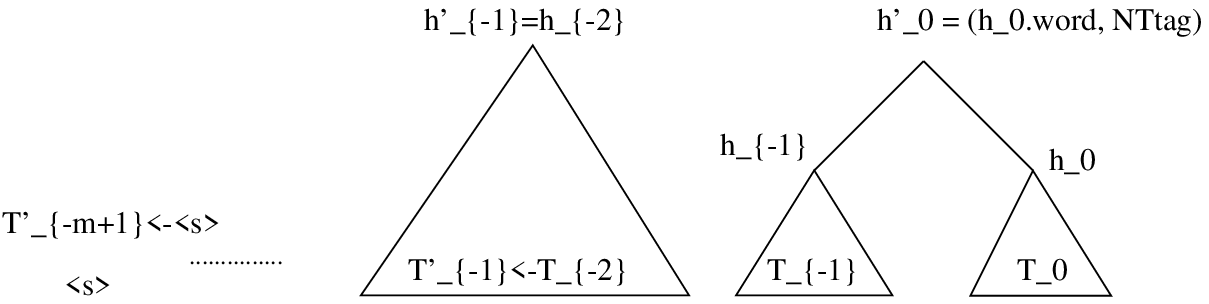,width=8cm}
    \caption{Result of adjoin-right under NTtag} \label{fig:after_a_r}
  \end{center}
\end{figure}

The operations performed by the PARSER are illustrated in
Figures~\ref{fig:before}-\ref{fig:after_a_r} and they ensure that all possible binary
branching parses with all possible headword and non-terminal tag
assignments for the $w_1 \ldots w_k$ word sequence can be generated.
It is easy to see that any given word sequence with a possible parse
and headword annotation is generated by a unique sequence of model
actions. 

\subsection{PROBABILISTIC MODEL} \label{section:prob_model}

The probability $P(W,T)$ of a word sequence $W$ and a complete parse
$T$ can be broken into:
\begin{eqnarray}
\lefteqn{P(W,T)=} \nonumber \\ 
& \prod_{k=1}^{n+1}[P(w_k/W_{k-1}T_{k-1}) \cdot P(t_k/W_{k-1}T_{k-1}, w_k) \cdot
  \prod_{i=1}^{N_k}P(p_i^k/W_{k-1}T_{k-1},w_k,t_k,p_1^k\ldots
  p_{i-1}^k)]
\end{eqnarray}
where:
\begin{itemize}
\item $W_{k-1} T_{k-1}$ is the word-parse $(k-1)$-prefix
\item $w_k$ is the word predicted by WORD-PREDICTOR
\item $t_k$ is the tag assigned to $w_k$ by the TAGGER
\item $N_k - 1$ is the number of operations the PARSER executes at
  position $k$ of the input string before passing control to the
  WORD-PREDICTOR (the $N_k$-th operation at position k is the
  \verb+null+ transition); $N_k$ is a function of $T$
\item $p_i^k$ denotes the i-th PARSER operation carried out at
  position k in the word string;\\
  $p_{i}^k \in\{$
  \verb+(adjoin-left, NTtag)+$,$
  \verb+(adjoin-right, NTtag)+$\}, 1 \leq i < N_k$ ,\\
  $p_{i}^k = $\verb+null+$, i = N_k$
\end{itemize}

Each \mbox{$(W_{k-1} T_{k-1},w_k,t_k,p_1^k \ldots p_{i-1}^k)$} is a
valid word-parse k-prefix $W_k T_k$ at position $k$ in the sentence,
$i=\overline{1,N_k}$. 

To ensure a proper probabilistic model certain PARSER and
WORD-PREDICTOR probabilities must be given specific values:
\begin{itemize}
\item  $P($\verb+null+$/W_kT_k) = 1$, if
\verb+h_{-1}.word = <s>+ and \verb+h_{0}+ $\neq$ \verb+(</s>, TOP')+
--- that is, before predicting \verb+</s>+ --- ensures that \verb+(<s>, SB)+
is adjoined in the last step of the parsing process; 
\item  $P($\verb+(adjoin-right, TOP)+$/W_k T_k) =1$,\\
if \verb+h_0 = (</s>, TOP')+ and \verb+h_{-1}.word = <s>+ and\\
$P($\verb+(adjoin-right, TOP')+$/W_k T_k) =1$,\\
if \verb+h_0 = (</s>, TOP')+ and \verb+h_{-1}.word+ $\neq$ \verb+<s>+ 
ensure that the parse generated by our model is consistent with the
definition of a complete parse;
\item $\exists\epsilon > 0, \forall W_{k-1}T_{k-1},
  P(w_k$=\verb+</s>+$/W_{k-1}T_{k-1})\geq\epsilon$ ensures that the
model halts with probability one.
\end{itemize}

In order to be able to estimate the model components we need to make
appropriate equivalence classifications of the conditioning part for
each component, respectively. 

The equivalence classification should identify the strong predictors in
the context and allow reliable estimates from a treebank. Our choice
is inspired by~\cite{mike96}:
\begin{eqnarray}
  P(w_k/W_{k-1} T_{k-1}) =& P(w_k/[W_{k-1} T_{k-1}]) & = P(w_k/h_0, h_{-1}) \label{eq:word_predictor_prob}\\
  P(t_k/w_k,W_{k-1} T_{k-1}) =& P(t_k/w_k,[W_{k-1} T_{k-1}]) & = P(t_k/w_k, h_0.tag, h_{-1}.tag)\\
  P(p_i^k/W_{k}T_{k}) =& P(p_i^k/[W_{k}T_{k}]) & = P(p_i^k/h_0, h_{-1})
\end{eqnarray}

It is worth noting that if the binary branching structure
developed by the parser were always right-branching and we mapped the
POStag and non-terminal tag vocabularies to a single type then our
model would be equivalent to a trigram language model.

\subsection{SMOOTHING} \label{subsection:modeling_tools}

All model components --- WORD-PREDICTOR, TAGGER, PARSER --- are
conditional probabilistic models of the type $P(y/x_1, x_2, \ldots ,x_n)$ where $y,
x_1, x_2, \ldots ,x_n$ belong to a mixed bag of words, POStags,
non-terminal tags and parser operations ($y$ only). 
For simplicity, the smoothing method we chose was deleted
interpolation among relative frequency estimates of different orders $f_n(\cdot)$
using a recursive mixing scheme:
\begin{eqnarray}
\lefteqn{P(y/x_1,\ldots,x_n) =} \nonumber\\
&\lambda(x_1,\ldots,x_n) \cdot P(y/x_1,\ldots,x_{n-1}) + (1 - \lambda(x_1, \ldots ,x_n)) \cdot f_n(y/x_1, \ldots ,x_n),\\
\lefteqn{f_{-1}(y) = uniform(vocabulary(y))}
\end{eqnarray}
The $\lambda$ coefficients are tied based on the range into which the count
$C(x_1, \ldots , x_n)$ falls.  The approach is a standard one~\cite{jelinek80}.

\subsection{PRUNING STRATEGY}

Since the number of parses  for a given word prefix $W_{k}$ grows
exponentially with $k$, $|\{T_{k}\}| \sim O(2^k)$, the state space of
our model is huge even for relatively short sentences. We thus have to 
prune most parses without discarding the most likely ones for a given
sentence $W$.  Our pruning strategy is a synchronous multi-stack
search algorithm.

Each stack contains hypotheses --- partial parses --- that have
been constructed by \emph{the same number of predictor and the same number of parser
operations}. The hypotheses in each stack are ranked according to the
$\ln(P(W_k,T_k))$ score, highest on top. 
The width of the search is controlled by two parameters:
\begin{itemize}
\item the maximum stack depth --- the maximum number of hypotheses
  the stack can contain at any given time; 
\item log-probability threshold --- the difference between the log-probability score of the top-most
  hypothesis and the bottom-most hypothesis at any given state of the
  stack cannot be larger than a given threshold.
\end{itemize}

\subsection{WORD LEVEL PERPLEXITY}\label{section:word_level_ppl}

Attempting to calculate the conditional perplexity by assigning to a whole sentence
the probability:
\begin{eqnarray}
  P(W/T^*) = \prod_{k=0}^n P(w_{k+1}/W_{k}T_{k}^*), \label{eq:ppl0}
\end{eqnarray}
where $T^* = argmax_{T}P(W,T)$ --- the search for $T^*$ being carried according
to our pruning strategy --- is not valid
because it is not causal: when predicting $w_{k+1}$ we would be using $T^*$ which was
determined by looking at the entire sentence. To be able to compare 
the perplexity of our model with that resulting from the standard
trigram approach, we need to factor in the entropy of
guessing the prefix of the final best parse $T_{k}^*$ \emph{before predicting} $w_{k+1}$,
based solely on the word prefix $W_{k}$.

To maintain a left-to-right operation of the language model, the
probability assignment for the word at position $k+1$ in the input
sentence was made using:
\begin{eqnarray}
  P(w_{k+1}/W_{k})=\sum_{T_{k}\in S_{k}}P(w_{k+1}/W_{k}T_{k})\cdot\rho(W_{k},T_{k}),\label{eq:ppl1}\\
  \rho(W_{k},T_{k}) = P(W_{k}T_{k})/\sum_{T_{k} \in S_{k}}P(W_{k}T_{k})
\end{eqnarray}
where $S_{k}$ is the set of all parses present in our stacks at the current stage $k$.

Note that if we set $\rho(W_{k},T_{k}) = \delta(T_k, T_k^*|W_k)$ ---
0-entropy guess for the prefix of the parse $T_k$ to equal that of the
final best parse $T_k^*$--- the two probability assignments
(\ref{eq:ppl0}) and  (\ref{eq:ppl1}) would be the same, yielding a lower
bound on the perplexity achievable by our model when using a given
pruning strategy. 

A second important observation is that the next-word predictor
probability $P(w_{k+1}/W_{k}T_{k})$ in~(\ref{eq:ppl1}) 
{\it need not be the same} as the WORD-PREDICTOR
probability~(\ref{eq:word_predictor_prob}) used to extract the structure
$T_k$, thus leaving open the possibility to estimate it separately. 

\subsection{PARAMETER REESTIMATION}\label{section:model_reest}

\subsubsection{First Model Reestimation}\label{section:model_reest_first_pass}
Our parameter re-estimation is inspired by the usual EM approach. Let
$(W,T^{(k)}),k=1,2,\ldots,N$ denote the set of parses of $W$ that
survived our pruning strategy. Each parse was produced by a unique
sequence of model actions: predictor, tagger, and parser moves. The
collection of these moves will be called a {\it derivation. }Each of
the $N$ members of the set is produced by exactly the same number of
moves of each type. Each move is uniquely specified by identifiers
$(y^{(m)},\underline{x}^{(m)}),$ where $m\in \{$WORD-PREDICTOR,
TAGGER, PARSER$\}$ denotes the
particular model, $y^{(m)}$ is the specification of the particular move
taken (e.g., for $m= $PARSER, the quantity $y^{(m)}$ specifies a choice from
$\{left,right,null\}$ and the exact tag attached), and
$\underline{x}^{(m)}$ specifies the move's context (e.g., for $m= $PARSER,
the two heads).

For each possible value $(y^{(m)},\underline{x}^{(m)})$ we will establish a counter
which at the beginning of any particular iteration will be empty. For each
move $(y^{(m)},\underline{x}^{(m)})$ present in the derivation of $(W,T^{(j)})$ we
add to the counter specified by $(y^{(m)},\underline{x}^{(m)})$ the amount 
$$
\rho (W,T^{(k)})=\frac{P(W,T^{(k)})}{\sum_{j=1}^NP(W,T^{(j)})} 
$$
where $P(W,T^{(j)})$ are evaluated on the basis of the model's parameter
values established at the end of the preceding iteration. We do that for all 
$(W,T^{(j)}),j=1,2,\ldots,N$ and for all sentences $W$ in the training data.
Let $C^{(m)}(y^{(m)},\underline{x}^{(m)})$ be the counter contents at the end of this
process. The corresponding relative frequency estimate will be%
$$
f(y^{(m)}|\underline{x}^{(m)})=\frac{C^{(m)}(y^{(m)},\underline{x}^{(m)})}{\sum_{z^{(m)}}C^{(m)}(z^{(m)},%
\underline{x}^{(m)})} 
$$
The lower order frequencies needed for the deleted interpolation of
probabilities in the next iteration are derived in the obvious way from the
same counters.

It is worth noting that because of pruning (which is a function of the
statistical parameters in use), the sets of surviving parses
$(W,T^{(k)}),k=1,2,\ldots,N$ for the same sentence $W$ may be
completely different for different iterations. 

\subsubsection{First Pass Initial Parameters}\label{section:initial_parameters}

Each model component --- WORD-PREDICTOR, TAGGER, PARSER ---
is initialised from a set of hand-parsed sentences, after
each parse tree $(W,T)$ is decomposed into its $derivation(W,T)$.
Separately for each $m$ model component, we:
\begin{itemize}
\item gather joint counts $C^{(m)}(y^{(m)}, {\underline{x}}^{(m)})$ from the derivations that
make up the ``development data'' using $\rho(W,T) = 1$;
\item estimate the deleted interpolation coefficients on joint
counts gathered from ``check data'' using the EM algorithm~\cite{em77}.
These are the initial parameters used with the reestimation procedure
described in the previous section.
\end{itemize}

\subsubsection{Language Model Refinement}\label{section:model_reest_second_pass}

In order to improve performance, we develop a model to be used
in~(\ref{eq:ppl1}), different from the  WORD-PREDICTOR
model~(\ref{eq:word_predictor_prob}). We will call this new component the
L2R-WORD-PREDICTOR.

The key step is to recognize in~(\ref{eq:ppl1}) a hidden Markov model
(HMM) with fixed transition probabilities --- although dependent on the
position in the input sentence $k$ --- specified by the
$\rho(W_{k},T_{k})$ values. 

The Expectation-step of the  EM algorithm~\cite{em77} for gathering
joint counts  $C^{(m)}(y^{(m)}, {\underline{x}}^{(m)})$, $m = $
L2R-WORD-PREDICTOR-MODEL, is the standard one whereas the
Maximization-step uses the same count smoothing technique as that
descibed in section~\ref{section:model_reest_first_pass}. 

The second reestimation pass is seeded with the $m = $ WORD-PREDICTOR model
joint counts  $C^{(m)}(y^{(m)}, {\underline{x}}^{(m)})$ resulting from
the first parameter reestimation pass (see section~\ref{section:model_reest_first_pass}).

\section{EXPERIMENTS} \label{section:experiments}

We have carried out the reestimation technique
described in section~\ref{section:model_reest} on 1 Mwds
of ``development'' data. For convenience we chose to work on the UPenn
Treebank corpus~\cite{Upenn} --- a subset of the WSJ (Wall Stree Journal) corpus. 
The vocabulary sizes were:\\
word vocabulary: 10k, open --- all words outside the vocabulary are
mapped to the \verb+<unk>+ token; POS tag vocabulary: 40, closed;
non-terminal tag vocabulary: 52, closed; parser operation vocabulary:
107, closed. The development set size was 929,564wds (sections 00-20), check set size
73,760wds (sections 21-22), test set size 82,430wds (sections 23-24).

Table~\ref{table:reest_ppls} shows the results of the reestimation techniques
presented in section~\ref{section:model_reest}; \verb+E?+ and
\verb+L2R?+ denote
iterations of the reestimation procedure described in sections~\ref{section:model_reest_first_pass} and \ref{section:model_reest_second_pass}, respectively.
A deleted interpolation  trigram model had perplexity 167.14 on the same
training-test data.
\begin{center}
  \begin{table}[h]
     \begin{center}
       \begin{tabular}{||l|c|c||} \hline
         iteration & DEV set & TEST set      \\ 
         number    & L2R-PPL & L2R-PPL       \\ \hline \hline
         E0        & 24.70   &  167.47       \\ \hline 
         E1        & 22.34   &  160.76       \\ \hline 
         E2        & 21.69   &  158.97       \\ \hline 
         E3 = L2R0 & 21.26   &  158.28       \\ \hline 
         L2R5      & 17.44   &  153.76       \\ \hline
       \end{tabular}
     \end{center}
     \caption{Parameter reestimation results} \label{table:reest_ppls}
   \end{table}
\end{center}
Simple linear interpolation between our model and the trigram model:
\begin{eqnarray*}
Q(w_{k+1}/W_k) = \lambda \cdot P(w_{k+1}/w_{k-1}, w_k) + (1 - \lambda) \cdot P(w_{k+1}/W_k) 
\end{eqnarray*} 
yielded a further improvement in PPL, as shown in
Table~\ref{table:interpolated_PPL}. The interpolation weight was
estimated on check data to be $\lambda = 0.36$. An overall relative
reduction of 11\% over the trigram model has been achieved.
\begin{center}
  \begin{table}[h]
     \begin{center}
       \begin{tabular}{||l|c|c||} \hline
         iteration & TEST set & TEST set\\ 
         number    & L2R-PPL  & 3-gram interpolated PPL   \\ \hline \hline
         E0        &  167.47  & 152.25  \\ \hline 
         E3        &  158.28  & 148.90  \\ \hline 
         L2R5      &  153.76  & 147.70  \\ \hline
       \end{tabular}
     \end{center}
     \caption{Interpolation with trigram results} \label{table:interpolated_PPL}
   \end{table}
\end{center}
As outlined in section~\ref{section:word_level_ppl}, the perplexity
value calculated using~(\ref{eq:ppl0}) is a lower bound for
 the achievable perplexity of our model; for the above search parameters and E3
 model statistics this bound was 99.60, corresponding to a relative reduction of
 41\% over the trigram model.

\section{CONCLUSIONS AND FUTURE DIRECTIONS} \label{section:future_work}

A new source model that organizes the prefix hierarchically in order
to predict the next symbol is developed. As a case study we applied
the source model to natural language, thus developing a new language
model with applicability in speech recognition. 

We believe that the above experiments show the potential
of our approach for improved language modeling for speech
recognition. Our future plans include:
\begin{itemize}
\item experiment with other parameterizations for the word predictor and parser models;
\item evaluate model performance as part of an automatic speech
  recognizer (measure word error rate improvement).
\end{itemize}

\section{Acknowledgments}
This research has been funded by the NSF IRI-19618874 grant (STIMULATE).

The authors would like to thank to Sanjeev Khudanpur for his
insightful suggestions. Also to Harry Printz, Eric Ristad, Andreas
Stolcke, Dekai Wu and all the other members of the dependency modeling
group at the summer96 DoD Workshop for useful comments on the model,
programming support and an extremely creative environment. Also thanks
to Eric Brill, Sanjeev Khudanpur, David Yarowsky, Radu Florian, Lidia
Mangu and Jun Wu for useful input during the meetings of the people
working on our STIMULATE grant.

\bibliographystyle{plain}
\bibliography{submission}
\end{document}